%% file: main.tex
\documentclass[10pt,twocolumn,letterpaper]{article}

\usepackage[pagenumbers]{iccv} %

\input{preamble}
\definecolor{iccvblue}{rgb}{0.21,0.49,0.74}
\usepackage[pagebackref,breaklinks,colorlinks,allcolors=iccvblue]{hyperref}
\usepackage{multirow}
\def\paperID{*****} %
\def\confName{ICCV}
\def\confYear{2025}

\title{FairReason: Balancing Reasoning and Social Bias in MLLMs}

\author{Zhenyu Pan$^{1}$, Yutong Zhang$^{2}$, Jianshu Zhang$^{1}$, Haoran Lu$^{1}$, Haozheng Luo$^{1}$, Yuwei Han$^{2}$ \\ Philip S. Yu$^{2}$, Manling Li$^{1}$, Han Liu$^{1}$\\ 
$^{1}$\textbf{Northwestern University}~~~ $^{2}$\textbf{University of Illinois at Chicago}
}
\begin{document}
\maketitle
\input{sec/0_abstract}    
\input{sec/1_intro}

\input{sec/2_relatedwork}
\input{sec/3_experiment}
\input{sec/4_result}
\input{sec/5_conclusion}
{
    \small
    \bibliographystyle{ieeenat_fullname}
    \bibliography{main}
}

\input{sec/X_suppl}

\end{document}

%% file: sec/0_abstract.tex
\begin{abstract}

Multimodal Large Language Models (MLLMs) already achieve state‑of‑the‑art results across a wide range of tasks and modalities. To push their reasoning ability further, recent studies explore advanced prompting schemes and post‑training fine‑tuning. Although these techniques improve logical accuracy, they frequently leave the models’ outputs burdened with pronounced social biases. Clarifying how reasoning gains interact with bias mitigation—and whether the two objectives inherently trade off—therefore remains an open and pressing research problem. Our study begins by benchmarking three bias‑mitigation strategies—supervised fine‑tuning (SFT), knowledge distillation (KD), and rule‑based reinforcement learning (RL)—under identical conditions, establishing their baseline strengths and weaknesses. Building on these results, we vary the proportion of debias-focused and reasoning-centric samples within each paradigm to chart the reasoning-versus-bias trade-off. Our sweeps reveal a consistent sweet spot: a roughly 1:4 mix trained with reinforcement learning cuts stereotype scores by 10\% while retaining 88\% of the model’s original reasoning accuracy, offering concrete guidance for balancing fairness and capability in MLLMs.
\end{abstract}

%% file: sec/1_intro.tex
\section{Introduction}
\label{sec:intro}

MLLMs perform well across various applications, including question answering \cite{pan2024chain, pan2024conv}, code generation \cite{li2024mmcode, pan2025code,pan2024codev}, and task automation \cite{shao2025sciscigpt}. To further improve their reasoning capabilities, recent works propose different methods, such as post-training fine-tuning \cite{thawakar2025llamavo1,pan2025metaspatial,huang2025visionr1incentivizingreasoningcapability}. However, although these methods raise benchmark scores, they neglect to consider the biases that appear in their generated outputs—biases inherited from the training data. Understanding how reasoning improvements interact with bias mitigation, and whether the two objectives inherently trade off, remains an important question for our community.

While previous studies suggest that reasoning improvements may support bias mitigation \cite{kabra2025reasoningfairnessmitigatingbias, wu2025doesreasoningintroducebias}, we revisit this assumption and find that it does not consistently hold, particularly for small-scale models trained with limited budgets. Our analysis reveals that this interaction highly depends on factors such as model size, training strategy, and the composition of training data, highlighting the need for a more nuanced and context-aware understanding of fairness in MLLMs. To better understand this dynamic, we conduct this systematic empirical study across multiple model architectures and training paradigms. Through this study, we offer new insights into how reasoning and fairness can be jointly optimized and point toward practical strategies for achieving better trade-offs in resource-constrained settings.

\begin{figure}
  \centering
  \includegraphics[width=0.4\textwidth]{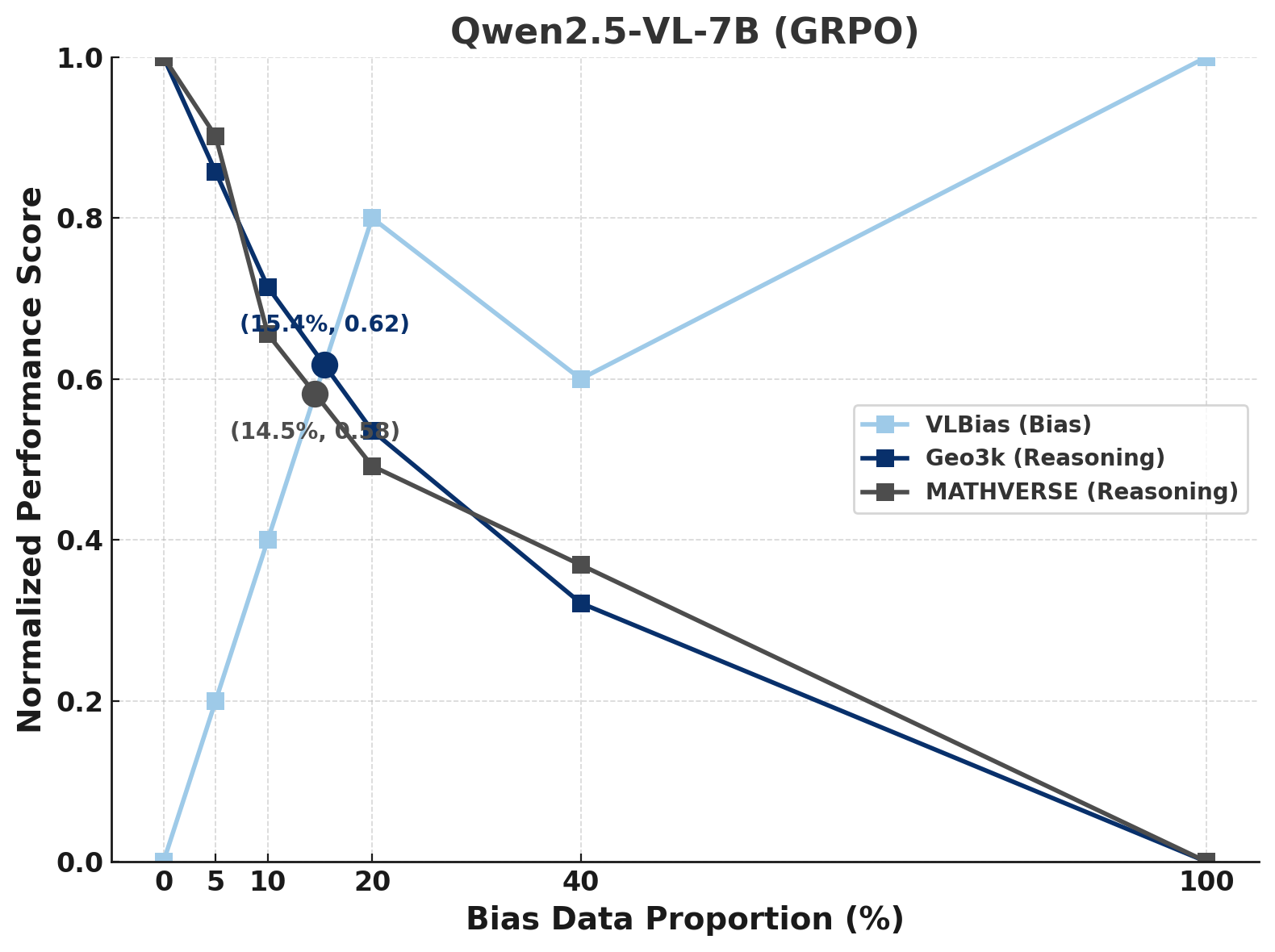}
  \caption{Qwen2.5-vl-7B sweet spot between reasoning and bias}
  \label{fig:beststrategy}
  \vspace{-0.2in}
\end{figure}

In the first stage of our study, we benchmark three bias mitigation strategies—supervised fine-tuning, knowledge distillation, and reinforcement learning-based methods—under consistent settings. Our results show that reinforcement learning yields superior performance compared to other training strategies, striking a better bias mitigation.
Building on this, the second stage explores how to balance reasoning and fairness by varying the composition of debias-oriented and reasoning-oriented training data. For each training paradigm, we identify data configurations that optimize both objectives. Our experiments reveal a consistent “sweet spot” in data distribution (e.g., 1:4 ratio) that significantly reduces bias without compromising reasoning accuracy. We release our best-performing models on Hugging Face to facilitate future research in fair and capable MLLMs.

In summary, our contributions are threefold: (1) we explore the dynamic relationship between reasoning and bias mitigation in MLLMs, showing that improvements in reasoning do not necessarily lead to fairer outputs; (2) we benchmark three training paradigms—supervised fine-tuning (SFT), knowledge distillation (KD), and rule-based reinforcement learning (RL)—for their effectiveness in reducing bias while preserving reasoning ability; and (3) we identify a consistent “sweet spot” in data composition for balancing reasoning and fairness under limited training budgets, and release our best-performing models—trained with these configurations—on Hugging Face to support reproducibility and future research.

%% file: sec/2_relatedwork.tex
\section{Related Work}
\label{sec:relatedwork}
We first introduce recent advances in MLLMs' reasoning and bias mitigation separately. We then review the emerging studies that explore how these two lines of work intersect.

\subsection{Reasoning in MLLMs}
Researchers propose various approaches to enhance the reasoning capabilities of MLLMs. One prominent direction is instruction tuning and fine-tuning on reasoning-focused datasets, which aim to strengthen logical and mathematical reasoning skills. Representative works include Math-LLaVA\cite{shi2024mathllavabootstrappingmathematicalreasoning}, LlamaV-o1\cite{thawakar2025llamavo1}, and Vision-R1\cite{huang2025visionr1incentivizingreasoningcapability}. Beyond supervised fine-tuning, recent efforts also explore reinforcement learning-based techniques such as Bootstrapped Preference Optimization (BPO)\cite{pi2024strengtheningmultimodallargelanguage} and Group Relative Policy Optimization (GRPO)\cite{huang2025visionr1incentivizingreasoningcapability}, which further incentivize multi-step reasoning through reward-driven feedback. These methods demonstrate improved performance on reasoning benchmarks, yet often overlook the fairness or bias implications of enhanced reasoning.

\subsection{Bias Mitigation in MLLMs}

MLLMs always exhibit social biases, reflecting and amplifying societal stereotypes present in their multimodal training data\cite{zhou-etal-2022-vlstereoset}. Mitigating these biases is challenging due to their complex architecture and the diverse sources of bias across both textual and visual modalities. To address this, researchers propose various strategies. From the data perspective, approaches such as dataset reweighting and targeted augmentation aim to diversify training distributions and reduce stereotypical associations, particularly those related to gender and race \cite{alabdulmohsin2024clipbiasusefulbalancing}. On the model level, adversarial debiasing techniques use auxiliary models to suppress biased representations, though often at the cost of performance \cite{berg-etal-2022-prompt}. Reinforcement Learning is also employed to encourage ethical alignment by penalizing biased outputs, albeit sometimes with utility trade-offs. In addition, post-hoc methods—such as output filtering, reranking \cite{zhang2024benchmarking}, and localized model editing \cite{10.1145/3664647.3681589}—seek to refine generated outputs without modifying the model's parameters. While these approaches show promise, they often focus on output control or representation adjustment, leaving open questions about how reasoning capabilities and bias mitigation interact within MLLMs.
\subsection{Reasoning with Bias in Language Models}

A few studies leverage reasoning to improve fairness in language models, but they primarily focus on proposing specific methods rather than analyzing the underlying relationship between reasoning and bias. For instance, reasoning-guided fine-tuning \cite{kabra2025reasoningfairnessmitigatingbias}, Bias-Augmented Consistency Training (BCT) \cite{chua2025biasaugmentedconsistencytrainingreduces}, and logical validation chains for stereotype detection \cite{tian2024rolereasoningidentificationsubtle} all illustrate that reasoning can aid in bias mitigation. However, these works stop short of offering a systematic investigation into the interplay between reasoning and fairness, and do not examine how this relationship varies across model sizes, training paradigms, or data compositions. Understanding this dynamic remains an open and underexplored challenge.

%% file: sec/3_experiment.tex
\section{Experiment Design}
\label{sec:experiment}

This section details the experimental setup of the empirical study. We investigate the impact of training data category and distribution on the trade-off between reasoning performance and social bias mitigation in MLLMs. We describe our research questions formally in the next subsection.

\subsection{Research Questions}

We aim to answer the following research questions:

\begin{itemize}
\item \textbf{Question 1}: Which training strategy is the most effective in mitigating generational social bias in LLMs and MLLMs?
\item \textbf{Question 2}: Under a fixed data budget, what proportion of reasoning-centric versus bias-centric data achieves the optimal trade-off between reasoning and bias mitigation across different training paradigms for both LLMs and MLLMs?
\end{itemize}

\subsection{Model}

We select two MLLM families, Qwen2.5-VL \cite{bai2025qwen25vltechnicalreport} and InternVL3 \cite{zhu2025internvl3exploringadvancedtraining}. These families demonstrate strong performance across modalities and tasks. We include the Qwen3 model family \cite{qwen3technicalreport} in our experiments to broaden our analysis to LLMs and make our results more generalizable.

\subsection{Datasets}

We utilize the Mix of Thoughts dataset~\cite{openr1} for LLMs and LLaVA-CoT-100k \cite{xu2024llavacot} for training MLLMs. The first dataset was created by researchers to reproduce results from the DeepSeek distilled model and to achieve comparable performance with distilled models from DeepSeek. The second dataset was created to train MLLMs that can reason in vision. For our experiments, we use subsets of the two datasets, consisting of approximately 5,000 samples each, to ensure the quality of model distillation, supervised fine-tuning, and GRPO. We select only 5,000 samples to demonstrate the relative balance between social bias mitigation and reasoning capabilities, rather than aiming for state-of-the-art (SOTA) results in MLLM or LLM reasoning.

For distillation training with high-quality reasoning supervision, we extract reasoning traces using two SOTA reasoning models: DeepSeek-R1 \cite{deepseekai2025deepseekr1incentivizingreasoningcapability} and OpenAI’s o4-mini. Since DeepSeek-R1 does not support multimodal inputs, we leverage it to generate reasoning traces for unimodal (text-only) datasets, while o4-mini is used for datasets involving multimodal reasoning. These two models are among the strongest available for reasoning tasks, making them well-suited to serve as teacher models in our distillation framework. We craft prompts to elicit both step-by-step reasoning traces and final answers for questions drawn from two benchmarks: the BBQ benchmark \cite{parrish-etal-2022-bbq} and VLBiasBench \cite{wang2024vlbiasbench}. For BBQ, we randomly sample training examples across all categories; for VLBiasBench, we focus on the closed-ended samples from its base section. In total, we collect approximately 3,000 reasoning traces per dataset to serve as supervision signals during training.
\subsection{Experiment Setup}

We evaluate three training schemes for bias mitigation to address our first research question: (1) supervised fine-tuning, (2) distillation from models, and (3) RL-based Group Relative Policy Optimization (GRPO). We compare these paradigms across both LLMs and MLLMs: Qwen3-8B, Qwen2.5-VL-7B, and InternVL3-8B. We utilize a fixed sample of 3k entries from the BBQ and VLBiasBench datasets for model training and use the same sampling strategies to ensure fairness in our comparisons. We provide training parameters, sampling strategies, and evaluation prompts in Appendix~\ref{sec:supp}.

To address the second research question, we explore data distribution strategies that balance reasoning capabilities and bias mitigation performance. We focus on two training schemes: distillation from models and GRPO. Supervised fine-tuning exhibits good performance in both reasoning enhancement and bias mitigation, but its training mechanics are the same as those of distillation, so we discard it here. We investigate proportions of reasoning data for each training scheme: 5\%, 10\%, 20\%, and 40\%. We evaluate the trained models' performance across benchmarks for both bias and reasoning capabilities to ensure that the models can achieve a balance between bias mitigation and reasoning. 

\subsection{Benchmarks}

We select benchmarks that reflect both bias mitigation and reasoning capabilities in LLMs and MLLMs. We employ two benchmarks for bias mitigation evaluation: the BBQ Benchmark \cite{parrish-etal-2022-bbq}, a multiple-choice question-answering dataset that measures social biases in language models; and VLBiasBench \cite{wang2024vlbiasbench}, a multimodal benchmark that assesses biases across nine social categories in vision-language models. We utilize four benchmarks for reasoning ability evaluation: AIME 2024, which includes challenging problems from the American Invitational Mathematics Examination (AIME); MATH-500 \cite{lightman2023lets}, a subset of 500 competition-level math problems across domains like algebra and geometry; MathVerse \cite{zhang2024mathverse}, a visual math benchmark designed to evaluate the multi-modal mathematical reasoning skills of MLLMs, focusing on their ability to interpret diagrams in visual math problems; and Geometry-3K \cite{lu2021inter}, a large-scale dataset comprising 3,002 multiple-choice geometry problems with dense annotations in formal language for diagrams and text, aimed at assessing geometry problem-solving capabilities.

%% file: sec/4_result.tex
\section{Empirical Findings}
\label{sec:result}

In this section, we briefly present our experimental results and provide a concise overview of the insights we gained throughout our study, which inform our answers.

\subsection{What is the best training strategy for bias mitigation}

\begin{figure*}
  \centering
  \includegraphics[width=1\textwidth]{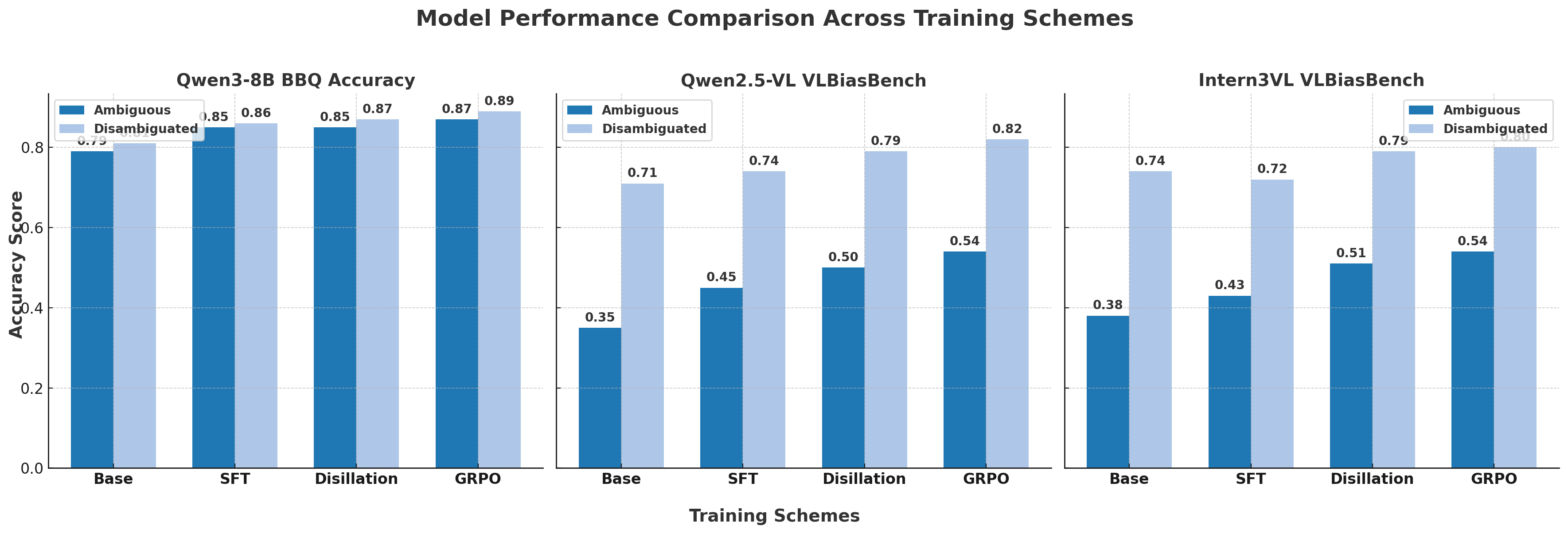}
  \caption{\textbf{Model Performance Comparison Across Training Schemes}.GRPO consistently improves bias mitigation, boosting Qwen3-8B’s ambiguous score from 0.79 to 0.87 on BBQ, and raising Qwen2.5-VL and Intern3VL from 0.35/0.38 to 0.54 on VLBiasBench, outperforming SFT and Distillation.}
  \label{fig:beststrategy}
\end{figure*}

\begin{figure*}
  \centering
  \includegraphics[width=1\textwidth]{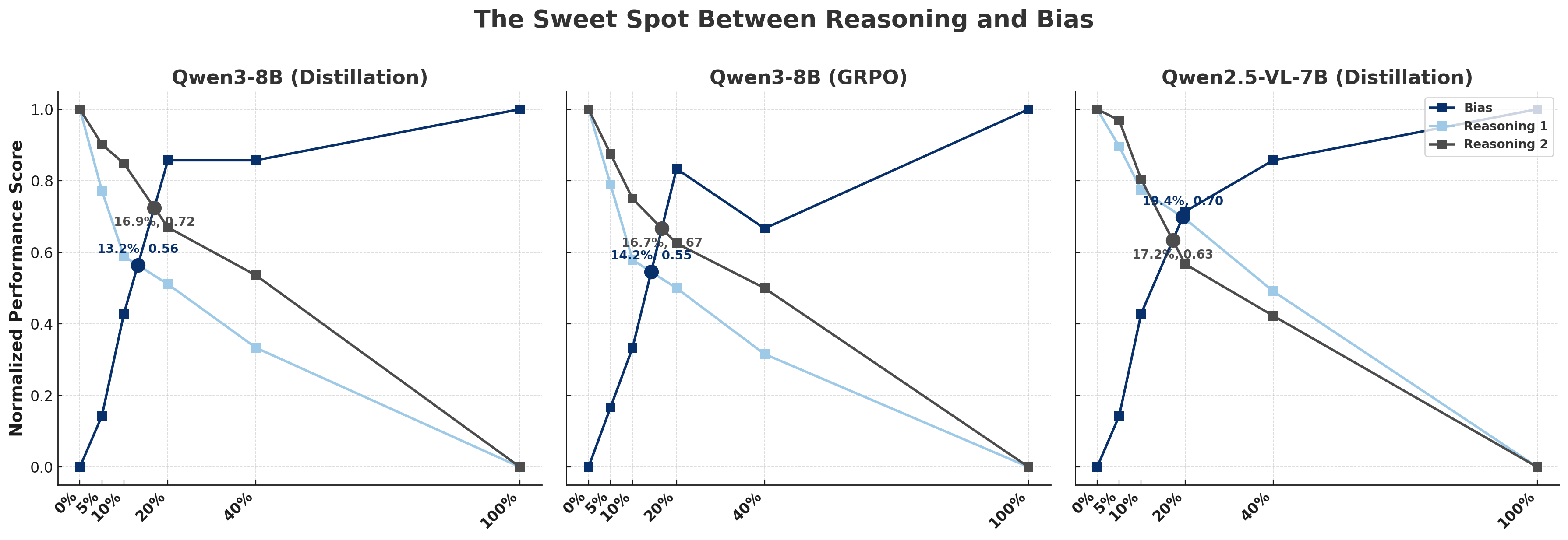}
  \caption{\textbf{The Sweet Spot Between Reasoning and Bias.}. Varying the ratio of debiasing data reveals a consistent trade-off curve. Across models, 10–20\% debiasing yields the best balance—e.g., GRPO on Qwen3-8B reduces bias by 14.2\% with minimal reasoning loss.}
  \label{fig:sweetspot}
\end{figure*}

We employ three training strategies to train three different models and evaluate their performance on different subsets of the training data. For the BBQ benchmark, we use a subset of 5k data from the original dataset, and for VLBiasBench, we use another 5k data from the base scene in the closed-ended questions. For both benchmarks, we evaluate the models' performance in both ambiguous and disambiguated scenes. We present the results of training LLMs and MLLMs for bias mitigation using different training strategies in Figure \ref{fig:beststrategy}. Across all model families, we found that the reinforcement learning-based method performs the best across all three training schemes and among all scenarios. This phenomenon can be attributed to the model having more freedom to explore ways to reduce bias in generation.

\subsection{The best data distribution for balancing reasoning and bias mitigation}

We try to find the best data mix for two kinds of training strategies (Model Distillation, GRPO). We present the picture for comparing all the data distributions that we experiment with in \ref{fig:sweetspot}. Through experiments, we find that with balanced mixtures of debiased and reasoning-oriented datasets, we can achieve significantly improved performance of bias mitigation capabilities and minor degeneration in reasoning.

After evaluating our trained models with the same scaling strategies, we normalize the test results and plot them in the diagram. We can find from the plot that the best data proportion for both LLMs and MLLMs to strike a balance between reasoning and bias mitigation is around 20\% of the total data samples being bias-centric. Beyond 20\%, further increases in bias‑centric data yield diminishing returns on bias benchmarks but accelerate reasoning decline: moving from 20\% to 100\% bias adds only a few points of accuracy on BBQ/VLBiasBench while having a great degeneration on reasoning tasks.  We make a full table of test results in the appendix~\ref{sec:supp}.

%% file: sec/5_conclusion.tex
\section{Conclusion}
\label{sec:conclusion}

In this work, we investigate the trade-off between reasoning ability and bias mitigation in LLMs and MLLMs. Through a unified benchmarking of three training strategies—supervised fine-tuning, knowledge distillation, and reinforcement learning—we identify their respective strengths and limitations under controlled conditions. Notably, RL enables more flexible exploration, achieving stronger bias mitigation while preserving reasoning performance. By systematically varying the mix of debiasing and reasoning-focused training samples, we uncover a clear sweet spot: a 1:4 ratio under RL reduces stereotype scores by 10\% while maintaining 88\% of the original reasoning accuracy. Our findings offer practical insights into aligning fairness and capability in LLMs and MLLMs, and highlight the promise of RL-based approaches for socially responsible model development.

%% file: sec/X_suppl.tex
\clearpage
\setcounter{page}{1}
\maketitlesupplementary
\label{sec:supp}
\section{Training Hyperparameters}

In this section, we provide details of the framework and hyperparameter settings used for training. For SFT and model distillation, we utilize the \texttt{LLaMA-Factory} framework with hyperparameter configurations listed in Table \ref{tab:hyperparams_sft}. For GRPO, we utilize the \texttt{Easy-R1} framework with hyperparameter configurations listed in Table~\ref{tab:hyperparams_grpo}.

\begin{table}[h]
    \centering
    \renewcommand{\arraystretch}{1.2}
    \begin{tabular}{|l|l|}
        \hline
        \textbf{Parameter} & \textbf{Value} \\ \hline
        Lora Rank & 8 \\
        Lora Target & All \\
        Learning rate & $5 \times 10^{-4}$ \\ 
        Number of epochs & 1 \\ 
        Batch size for training & 1 \\ 
        Run validation & False \\ 
        Batching strategy & padding \\ 
        Context length & 10000 \\
        Gradient accumulation steps & 16 \\
        Gradient clipping & False \\ 
        Weight decay & 0.1 \\
        Seed & 42 \\
        Use FP16 precision & False \\
        Mixed precision & True \\ \hline
    \end{tabular}
    \caption{\textbf{Hyperparameter configurations used in SFT and Model Distillation.}}
    \label{tab:hyperparams_sft}
\end{table}

\begin{table}[h]
    \centering
    \renewcommand{\arraystretch}{1.2}
    \begin{tabular}{|l|l|}
        \hline
        \textbf{Parameter} & \textbf{Value} \\ \hline
        Learning rate & $1 \times 10^{-6}$ \\ 
        Number of epochs & 1 \\
        Batch size for training & 128 \\
        n & 5 \\
        Run validation & False \\ 
        Batching strategy & padding \\
        Context length & 2048 \\
        Gradient clipping & False \\ 
        Weight decay & $1 \times 10^{-2}$ \\
        Seed & 42 \\
        Use FP16 precision & False \\
        Mixed precision & True \\ \hline
    \end{tabular}
    \caption{\textbf{Hyperparameter configurations used in GRPO}}
    \label{tab:hyperparams_grpo}
\end{table}

\section{Test Results}
\label{sec:test}

We list the results for our test results in Table~\ref{tab:distillation} and Table~\ref{tab:grpo}.

\begin{table*}[ht]
\centering
\caption{Performance under Distillation Strategy}
\label{tab:distillation}
\begin{tabular}{lccccccc}
\toprule
Model             & Benchmark     & 0\%  & 5\%  & 10\% & 20\% & 40\% & 100\% \\
\midrule
\multirow{3}{*}{Qwen3‑8B} 
                  & BBQ           & 0.79 & 0.80 & 0.82 & 0.85 & 0.85 & 0.86 \\
                  & MATH          & 69.5 & 65.4 & 62.1 & 60.7 & 57.5 & 51.5 \\
                  & AIME\,2024    & 41.2 & 40.1 & 39.5 & 37.5 & 36.0 & 30.0 \\
\midrule
\multirow{3}{*}{Qwen2.5‑VL‑7B} 
                  & VLBiasBench   & 0.75 & 0.76 & 0.78 & 0.80 & 0.81 & 0.82 \\
                  & Geo3K         & 51.0 & 49.2 & 47.1 & 45.7 & 42.2 & 33.7 \\
                  & MATHVERSE     & 50.4 & 50.1 & 48.5 & 46.2 & 44.8 & 40.7 \\
\bottomrule
\end{tabular}
\end{table*}

\begin{table*}[ht]
\centering
\caption{Performance under GRPO Strategy (Scheme B)}
\label{tab:grpo}
\begin{tabular}{lccccccc}
\toprule
Model             & Benchmark     & 0\%  & 5\%  & 10\% & 20\% & 40\% & 100\% \\
\midrule
\multirow{3}{*}{Qwen3‑8B} 
                  & BBQ           & 0.82 & 0.83 & 0.84 & 0.87 & 0.86 & 0.88 \\
                  & MATH          & 71.0 & 67.0 & 63.0 & 61.5 & 58.0 & 52.0 \\
                  & AIME\,2024    & 43.0 & 41.5 & 40.0 & 38.5 & 37.0 & 31.0 \\
\midrule
\multirow{3}{*}{Qwen2.5‑VL‑7B} 
                  & VLBiasBench   & 0.78 & 0.79 & 0.80 & 0.82 & 0.81 & 0.83 \\
                  & Geo3K         & 70.0 & 68.0 & 66.0 & 63.5 & 60.5 & 56.0 \\
                  & MATHVERSE     & 53.2 & 52.0 & 49.0 & 47.0 & 45.5 & 41.0 \\
\bottomrule
\end{tabular}
\end{table*}

\section{Sampling Strategy}

For all the evaluations in our study, we use a random seed of 42, a maximum response token limit of 10,000, and generate 5 responses per prompt, with instructions for the models to enclose their final answers in \verb|\boxed{}|.